\documentclass[10pt, journal]{IEEEtran}
\usepackage{amsmath,amsfonts}
\usepackage{amssymb}
\usepackage{algorithmic}
\usepackage{algorithm}
\usepackage{array}
\usepackage{caption}
\usepackage{textcomp}
\usepackage{stfloats}
\usepackage{url}
\usepackage{graphicx}
\usepackage{booktabs}
\usepackage{cite}
\usepackage{xcolor}
\usepackage{tikz}
\usetikzlibrary{arrows.meta,backgrounds,calc,fit,positioning}
\usepackage{multirow}
\usepackage{booktabs}
\usepackage{tabularx}
\usepackage{makecell}
\usepackage{float}
\usepackage{threeparttable}
\usepackage[hidelinks]{hyperref}
\usepackage{cleveref}
\usepackage{microtype}

\newcommand{\eg}{\textit{e.g.,}}%

\crefname{figure}{Fig.}{Figs.}
\Crefname{figure}{Fig.}{Figs.}
\crefname{section}{Sec.}{Secs.}
\Crefname{section}{Section}{Sections}
\crefname{table}{Table}{Tables}
\Crefname{table}{Table}{Tables}

\newcolumntype{Y}{>{\raggedright\arraybackslash}X}

\makeatletter
\renewcommand{\thesection}{\arabic{section}}
\renewcommand{\thesubsection}{\thesection.\arabic{subsection}}
\renewcommand{\thesubsubsection}{\thesubsection.\arabic{subsubsection}}

\makeatletter
\renewcommand{\thesection}{\arabic{section}}
\renewcommand{\thesubsection}{\thesection.\arabic{subsection}}
\renewcommand{\thesubsubsection}{\thesubsection.\arabic{subsubsection}}

\renewcommand{\@seccntformat}[1]{%
  \csname the#1\endcsname.\enspace
}
\makeatother

\title{
    Occupancy-Grounded Room Segmentation for Hierarchical 3D Scene Graphs
}

\author{
    Carlos Cueto Zumaya*, Iacopo Catalano, Jorge Pe\~na-Queralta, and Wallace Moreira Bessa
    \thanks{* Corresponding author.}
    \thanks{Carlos Cueto Zumaya, Iacopo Catalano, and Wallace Moreira Bessa are with the University of Turku, 20014 Turku, Finland. (e-mail: \{crcuzu, imcata, wmobes\}@utu.fi).}
    \thanks{Jorge Pe\~na-Queralta is with the Centre for Artificial Intelligence, Z\"urich University of Applied Sciences, Winterthur, Switzerland. (e-mail: penq@zhaw.ch).}
}

\begin{document}

\maketitle

\begin{abstract}
Hierarchical 3D scene graphs (3DSGs) for indoor robots organize geometric and semantic information across spatial scales, with a room layer that connects object-level perception to room-scale reasoning. Existing systems construct this layer from different 
spatial substrates (\eg{} place clusters, wall planes, or segmentation outputs), and as a result, room nodes are not evaluated on a common geometric criterion. We present an occupancy-grounded 3DSG pipeline in which room nodes are anchored to tracked free-space 
regions derived from occupancy decomposition, giving each room an explicit polygonal footprint. We evaluate the pipeline on 12 Matterport3D scenes by matching predicted room polygons to annotated room instances and compare against Hydra, a representative state-of-the-art place-connectivity baseline. The results show that occupancy-grounded anchoring recovers substantially more room instances than place-connectivity construction, at the cost of lower precision, and that wall-accurate room boundaries remain an open problem for both methods. Code is available at \url{https://github.com/crcz25/OccuSG}.
\end{abstract}


\section{Introduction}
\label{sec:introduction}

Hierarchical 3D scene graphs (3DSGs) organize indoor environments as layered spatial-semantic representations, linking geometric observations to objects, traversable space, and room-scale structure~\cite{catalano20253d}. This representation %, richer than standard 3D maps, 
supports tasks that require reasoning across spatial scales, from locating objects to planning routes through adjacent spaces or interpreting room-level instructions. The room layer is central to this hierarchy, grouping free space and objects into the human-scale units that ground containment, room-conditioned reasoning, and inter-room navigation.

% Different systems construct this layer from the spatial substrate their mapping pipeline produces. Place-connectivity systems cluster free-space or GVD-based place nodes into rooms~\cite{rosinol2021kimera,hughes2022hydra,hughes2024foundations}.
Different approaches define the higher-level room layer in different ways. Place-connectivity methods typically group connected navigational nodes into rooms~\cite{rosinol2021kimera,hughes2022hydra,hughes2024foundations}. 
Others derive room-level entities from geometric wall planes~\cite{bavle2022situational,bavle2023s} or room-like regions from bird's-eye-view histograms and learned modules~\cite{werby2024hierarchical, xu2025point2graph}. 
However, these constructions are not equivalent, and none of them is evaluated on a direct geometric criterion that measures whether the predicted room nodes recover the physical room instances of the environment, their count, spatial extent, and boundaries.

We address this by grounding room construction in occupancy-derived free-space regions. The proposed pipeline builds a hierarchical 3DSG from RGB-D input by first collapsing a 3D occupancy map into a 2D free-space layer through vertical-clearance analysis~\cite{fredriksson2024voxel}, then partitioning that layer into polygonal regions via free-space decomposition~\cite{fermin2017incremental}, giving every room an explicit polygonal footprint.

Because room geometry is explicit, the room layer can be evaluated by matching predicted polygons to annotated room instances, a standard practice in the 2D room segmentation literature that has not previously been applied to 3DSG room layers.

\section{Related Work}
\label{sec:related_work}

Hierarchical 3DSGs have been used across a range of applications, including semantic mapping, task planning, and natural-language grounding~\cite{agia2022taskography,rana2023sayplan,maggio2024clio}, where room layers provide human-scale structure for reasoning and navigation. We review how room layers are constructed and why they vary, how they are evaluated, and what remains unmeasured in the literature.

% How room layers are constructed, and why they vary.
Room construction in 3DSG systems is not standardized; it is an emergent property of the spatial substrate each pipeline maintains.
Place-connectivity approaches group free-space nodes into rooms using topological cues and persistent homology~\cite{hughes2022hydra,hughes2024foundations}, so a predicted room corresponds to a cluster of navigable places rather than a delineated spatial region.
Structural approaches extract room-level entities from geometric wall planes~\cite{bavle2022situational,bavle2023s, bavle2025s}, tying room identity to architectural boundaries.
Segmentation-based and open-vocabulary approaches construct room-like entities from bird's-eye-view maps, volumetric boundaries, watershed grouping, or learned
modules~\cite{werby2024hierarchical,werby2025keysg,xu2025point2graph,
janzon2025enhancing}, so a predicted room reflects the output of a classifier or segmentation model rather than a geometric region.
These representations may all be useful within their respective pipelines, but they do not necessarily correspond to the same physical room instances.

% How room layers are evaluated, and what is left unmeasured.
Because room nodes are constructed differently across systems, their evaluation has remained tied to their respective pipelines, borrowing from related
tasks such as trajectory accuracy or semantic segmentation quality~\cite{wald2020learning,wu2021scenegraphfusion}. These metrics capture important
properties of the graph as a whole but measure components either below the room layer or above it, coupling their evaluation to the construction substrate~\cite{rosinol2021kimera, hughes2022hydra, maggio2024clio, werby2024hierarchical, xu2025point2graph}.
These evaluations are useful within their systems, but they do not provide a common basis for asking whether a predicted room node corresponds to a physical room instance in the environment.

% What occupancy-based segmentation offers.
In contrast to the above methods, occupancy-derived region segmentation offers a direct way to connect
room construction with the free-space structure of the environment. Unlike most 3DSG room layers, these regions can be evaluated against annotated room layouts with polygon-overlap metrics including IoU-based matching, precision, recall, F1, boundary F1, and over-/under-segmentation diagnostics~\cite{fermin2017incremental, luperto2022robust, jung2017automatic, hiller2019learning, bao2025topology}. These methods show that free-space structure can make room and region recovery measurable. However, they are typically developed and evaluated as stand-alone segmentation approaches rather than as mechanisms for generating room nodes inside a hierarchical 3DSG pipeline.

This leaves a gap between construction and evaluation
that we leverage to construct 
room-level graph nodes from occupancy-derived regions and evaluate them as 
spatial outputs matched to ground-truth room instances.

\section{Method}
\label{sec:method}
\begin{figure*}[ht]
\centering
\setlength{\abovecaptionskip}{2pt}
\setlength{\belowcaptionskip}{2pt}

\definecolor{sgInk}{RGB}{35,35,35}
\definecolor{sgGray}{RGB}{105,105,105}
\definecolor{sgLightGray}{RGB}{245,245,245}
\definecolor{sgBlue}{RGB}{0,114,178}
\definecolor{sgOrange}{RGB}{230,159,0}
\definecolor{sgGreen}{RGB}{0,158,115}
\definecolor{sgPurple}{RGB}{204,121,167}

\resizebox{0.96\textwidth}{!}{%
\begin{tikzpicture}[
  font=\small\sffamily,
  every node/.style={text=sgInk},
  panelTitle/.style={font=\small\sffamily, anchor=north, align=center},
  stagePurple/.style={draw=sgPurple!65!black, fill=sgPurple!10,
    line width=0.8pt, rounded corners=3pt},
  stageOrange/.style={draw=sgOrange!65!black, fill=sgOrange!10,
    line width=0.8pt, rounded corners=3pt},
  stageGreen/.style={draw=sgGreen!65!black,  fill=sgGreen!10,
    line width=0.8pt, rounded corners=3pt},
  mod/.style={draw=sgInk!50, fill=white, line width=0.5pt,
    rounded corners=2pt, align=center, font=\small\sffamily,
    minimum height=1.0cm},
  res/.style={draw=#1!75!black, fill=#1!22, line width=0.6pt,
    rounded corners=2pt, align=center, font=\small\sffamily,
    minimum height=1.0cm},
  inpbox/.style={draw=sgBlue!60!black, fill=sgBlue!8,
    rounded corners=3pt, line width=0.7pt},
  outbox/.style={draw=sgGray!70!black, fill=sgLightGray,
    rounded corners=3pt, line width=0.6pt},
  fl/.style={-Stealth, line width=0.85pt},
  dfl/.style={-Stealth, line width=0.85pt, dashed},
]

% =================== INPUT PANEL ==================================
\draw[inpbox] (-6.9, 0.2) rectangle (-1.0, 7.0);
\node[panelTitle] at (-3.95, 6.85) {Inputs};

% RGB-D
\draw[inpbox, fill=sgBlue!12] (-6.45, 4.55) rectangle (-1.45, 6.20);
\node[font=\small\sffamily, anchor=north] at (-3.95, 6.05)
  {RGB-D observations};
\node[font=\small\sffamily] at (-3.95, 5.35)
  {$I^{\mathrm{rgb}}_t,\; I^{\mathrm{depth}}_t$};
\node[font=\scriptsize\sffamily, text=sgGray] at (-3.95, 4.85)
  {color and depth frames};

% Odometry
\draw[inpbox, fill=sgBlue!12] (-5.95, 0.55) rectangle (-1.95, 1.95);
\node[font=\small\sffamily] at (-3.95, 1.60) {Odometry};
\node[font=\small\sffamily] at (-3.95, 1.03) {$T^{}_t$};

% =================== GEOMETRIC PERCEPTION (purple) ================
% Shortened slightly to leave a clean routing lane below it.
\draw[stagePurple] (-0.20, 1.25) rectangle (10.20, 7.0);
\node[panelTitle] at (5.00, 6.9)
  {Geometric perception~(\S\ref{subsec:low_level_graph})};

% Top lane — occupancy pipeline
\node[mod, minimum width=2.0cm] (pcg) at (1.05, 5.85)
  {Point cloud\\generation};
\node[mod, minimum width=2.0cm] (oct) at (3.45, 5.85)
  {3D occupancy\\octree};
\node[mod, minimum width=2.25cm] (fsp) at (6.10, 5.85)
  {2D free-space\\projection};
\node[res=sgPurple, text width=1.75cm] (fsl) at (8.95, 5.85)
  {Free-space\\layer $F^{}_t$};

\draw[fl, draw=sgPurple!80!black] (pcg.east) -- (oct.west);
\draw[fl, draw=sgPurple!80!black] (oct.east) -- (fsp.west);
\draw[fl, draw=sgPurple!80!black] (fsp.east) -- (fsl.west);

% Lane divider
\draw[sgGray!35, dashed, line width=0.4pt] (0.05, 3.85) -- (9.95, 3.85);

% Bottom lane — semantic detection pipeline, moved upward
\node[mod, text width=2.70cm] (yolo) at (1.45, 2.25)
  {YOLO segmentation\\(RGB $\to$ 2D masks)};
\node[mod, text width=2.85cm] (bkp) at (5.15, 2.25)
  {Depth backprojection\\(masks $+$ depth $\to$ 3D)};
\node[res=sgPurple, text width=1.75cm] (dtn) at (8.85, 2.25)
  {3D\\detections $D^{}_t$};

\draw[fl, draw=sgPurple!80!black] (yolo.east) -- (bkp.west);
\draw[fl, draw=sgPurple!80!black] (bkp.east) -- (dtn.west);

% =================== REGION GROUNDING (orange) ====================
\draw[stageOrange] (10.85, 3.9) rectangle (17.15, 7.0);
\node[panelTitle] at (14.00, 6.9)
  {Region grounding~(\S\ref{subsec:region_extraction})};

\node[mod, minimum width=2.2cm] (dude) at (12.20, 5.85)
  {DUDE\\decomp.};
\node[mod, minimum width=2.2cm] (rtrk) at (15.65, 5.85)
  {Region\\tracking};
\node[res=sgOrange, text width=3.35cm] (rtn) at (14.00, 4.55)
  {Tracked regions $R^{}_t$\\+ room anchoring};

% Explicit orthogonal routing points around the tracked-regions block
\coordinate (dudeDown) at (12.20, 5.20);
\coordinate (dudeSideTop) at (11.78, 5.20);
\coordinate (dudeSideIn) at (11.78, 4.55);

\coordinate (rtrkDown) at (15.65, 5.20);
\coordinate (rtrkSideTop) at (16.22, 5.20);
\coordinate (rtrkSideIn) at (16.22, 4.55);

% Main DUDE -> tracking flow
\draw[fl, draw=sgOrange!80!black] (dude.east) -- (rtrk.west);

% DUDE -> tracked regions
\draw[fl, draw=sgOrange!80!black]
  (dude.south) --
  (dudeDown) --
  (dudeSideTop) --
  (dudeSideIn) --
  (rtn.west);

% Region tracking -> tracked regions
\draw[fl, draw=sgOrange!80!black]
  (rtrk.south) --
  (rtrkDown) --
  (rtrkSideTop) --
  (rtrkSideIn) --
  (rtn.east);

% =================== GRAPH CONSTRUCTION (green) ==================
\draw[stageGreen] (10.85, 0.2) rectangle (17.15, 3.6);
\node[panelTitle] at (14.00, 3.5)
  {Graph construction~(\S\ref{subsec:graph_maintenance})};

\node[mod, text width=5.15cm, minimum height=2.5cm, align=left] (graphbody) at (14.00, 1.80) {%
  \hspace{0.5em}$\bullet$\enspace Room nodes \quad (from $R^{}_t$)\\[4pt]
  \hspace{0.5em}$\bullet$\enspace Object nodes\ (from $D^{}_t$)\\[4pt]
  \hspace{0.5em}$\bullet$\enspace Agent nodes \; (from $T^{}_t$)\\[4pt]
  \hspace{0.5em}$\bullet$\enspace Navigation nodes\ (from $F^{}_t$)%
};

% =================== OUTPUT IMAGE ONLY ============================
\node[inner sep=0pt] (sgimg) at (19.70, 3.50)
  {\includegraphics[width=3.45cm]{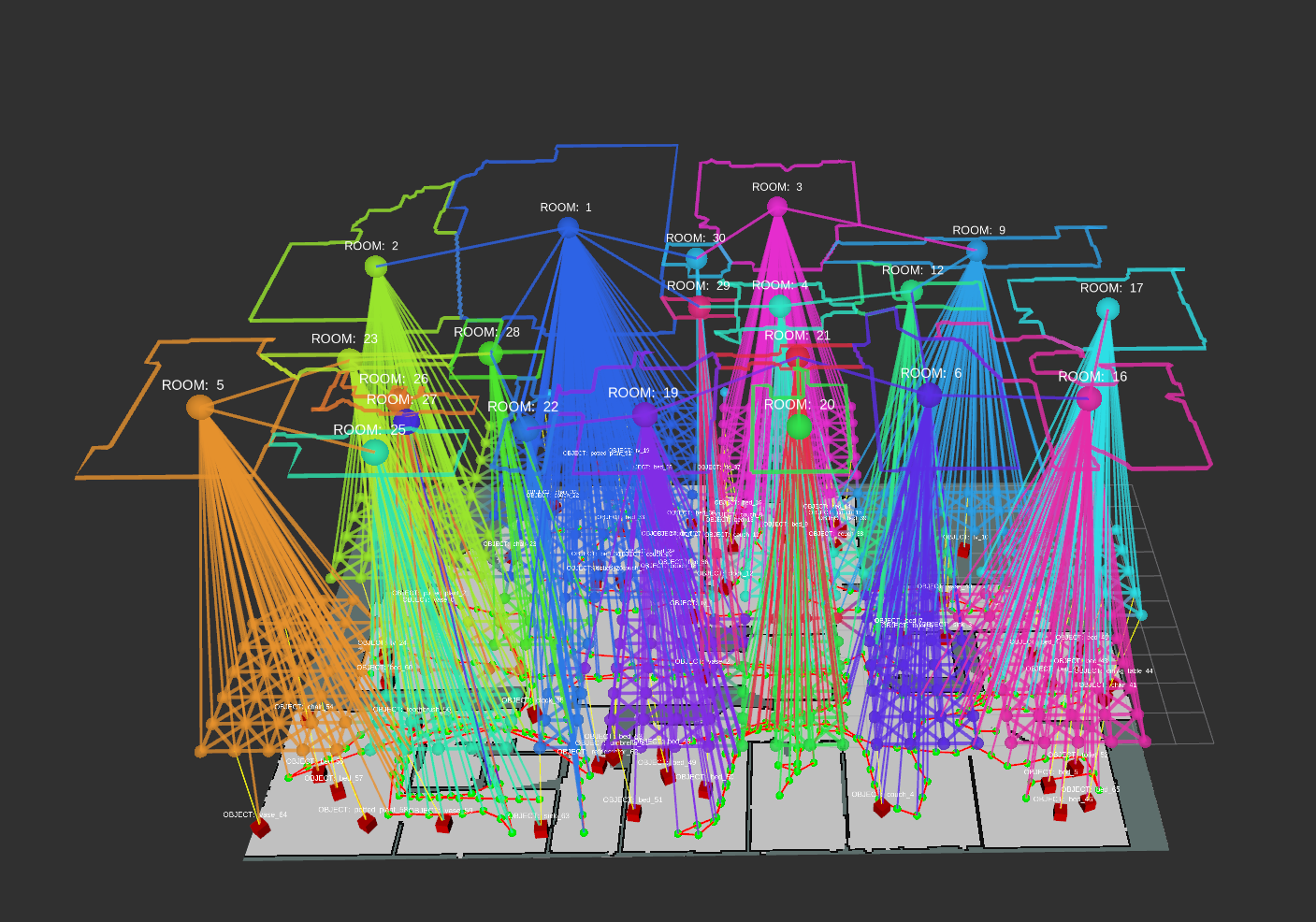}};

% Optional invisible coordinate to keep approximately the same right-side extent
\path (21.65, 7.0);

% =================== ARROWS ======================================

% RGB-D -> Point cloud generation
\draw[fl, draw=sgBlue!70!black]
  (-1.45, 5.85) -- (-0.55, 5.85) -- (pcg.west);

% RGB-D -> YOLO segmentation, adjusted to the raised semantic lane
\draw[fl, draw=sgBlue!70!black]
  (-1.45, 5.15) -- (-0.55, 5.15) --
  (-0.55, 2.25) -- (yolo.west);

% Odometry -> Graph construction, routed directly below the shortened purple panel
\draw[dfl, draw=sgGray]
  (-1.95, 0.85) -- (10.85, 0.85);

% Free-space layer -> Region grounding
\draw[fl, draw=sgOrange!80!black]
  (fsl.east) -- (dude.west);

% Free-space layer -> Graph construction
\coordinate (fsOut) at ($(fsl.east)!0.35!(fsl.south east)$);
\coordinate (fsGraphBusTop) at (10.45, 5.35);
\coordinate (fsGraphBusBot) at (10.45, 2.75);
\coordinate (fsGraphIn) at (10.85, 2.75);

\draw[dfl, draw=sgGreen!70!black]
  (fsOut) --
  (fsGraphBusTop) --
  (fsGraphBusBot) --
  (fsGraphIn);

% 3D detections -> Graph construction, now fully horizontal
\draw[dfl, draw=sgGreen!70!black]
  (dtn.east) -- (10.85, 2.25);

% Tracked regions -> Graph construction
\draw[fl, draw=sgGreen!80!black]
  (rtn.south) -- (14.00, 3.60);

% Graph construction -> Scene graph image
\draw[fl, draw=sgGray!70!black]
  (17.15, 1.80) -- (17.55, 1.80) --
  (17.55, 2.85) -- ([yshift=-0.65cm]sgimg.west);

% Region grounding -> Scene graph image
\draw[fl, draw=sgGray!70!black]
  (17.15, 5.45) -- (17.55, 5.45) --
  (17.55, 4.15) -- ([yshift=0.65cm]sgimg.west);

\end{tikzpicture}%
}
\caption{Pipeline overview.
\emph{Geometric perception} (\S\ref{subsec:low_level_graph}) fuses RGB-D frames
into a 3D occupancy map, projects it to a 2D free-space layer $F^{}_t$, and concurrently
runs YOLO segmentation with depth backprojection to produce 3D object
detections~$D^{}_t$.
\emph{Region grounding} (\S\ref{subsec:region_extraction}) decomposes the
free-space layer into polygonal regions via DUDE, tracks them over time, and
anchors one room node per stable region, yielding tracked regions~$R^{}_t$.
\emph{Graph construction} (\S\ref{subsec:graph_maintenance}) assembles all node
layers—room, object, agent, and navigation nodes—linked by containment and
traversability edges.
Solid arrows: primary geometric data flow; dashed arrows: secondary feeds into
graph assembly.}
\label{fig:scene_graph_pipeline}
\end{figure*}
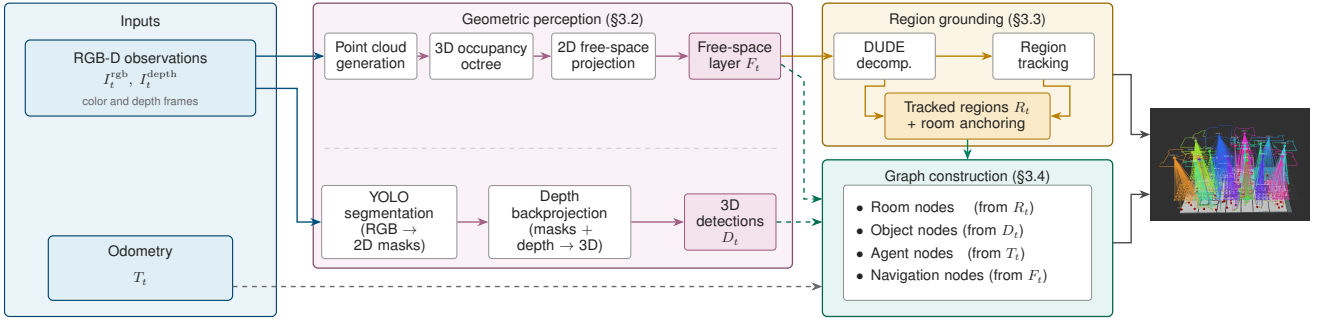

Our method constructs a hierarchical 3DSG whose room nodes are persistent, geometrically grounded entities with explicit polygonal footprints. As shown in \Cref{fig:scene_graph_pipeline}, RGB-D observations are integrated into an octree-based 3D occupancy map~\cite{hornung2013octomap}, projected into a 2D free-space map~\cite{fredriksson2024voxel}, decomposed into polygonal regions~\cite{fermin2017incremental}, and tracked over time to provide stable spatial supports for room nodes. Objects, robot poses, and navigation nodes are then assigned to the resulting room footprints, yielding a 3DSG whose room layer is topologically connected and directly evaluable against room-scale geometric annotations. 

\subsection{Hierarchical 3D Scene Graph}
\label{subsec:system_overview}
The scene graph $\mathcal{G}^{}_t = (\mathcal{V}^{}_t, \mathcal{E}^{}_t)$ contains four node layers: \textit{agent} nodes (sampled robot poses), \textit{object} nodes (merged 3D semantic detections), \textit{navigation} nodes (traversable free-space blocks), and \textit{room} nodes (hypotheses anchored to tracked free-space regions).
Edges encode relationships within and across layers: traversability between neighboring navigation nodes, containment of objects, adjacency between rooms sharing a navigable crossing, and temporal links between consecutive agent poses.

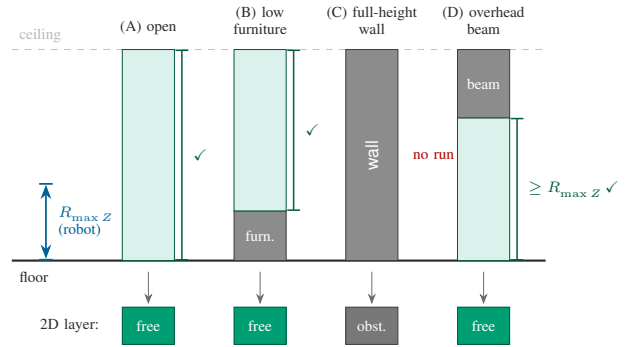
\begin{figure}[t]
\centering
\definecolor{fsFree}{RGB}{0,158,115}
\definecolor{fsOcc}{RGB}{120,120,120}
\definecolor{fsInk}{RGB}{45,45,45}
\definecolor{fsClr}{RGB}{0,114,178}

\resizebox{0.92\columnwidth}{!}{%
\begin{tikzpicture}[font=\sffamily\footnotesize, >=Stealth,
  occ/.style ={fill=fsOcc!85, draw=fsOcc!50!black, line width=0.4pt},
  free/.style={fill=fsFree!14, draw=fsFree!55!black, line width=0.4pt},
  lbl/.style ={fsInk, font=\scriptsize},
  clr/.style ={fsClr!85!black, line width=0.8pt},
  brace/.style={fsFree!60!black, line width=0.7pt}]

\def\hw{0.42}     % column half-width
\def\Hc{3.4}      % ceiling height
\def\Rz{1.25}     % required robot clearance R_maxZ

% ---- floor & ceiling -------------------------------------------------
\draw[fsInk, line width=1pt] (0.0,0) -- (8.6,0);
\node[lbl, anchor=north west] at (0.0,-0.05) {floor};
\draw[gray!50, dashed] (0.0,\Hc) -- (8.6,\Hc);
\node[lbl, anchor=south west, gray!55] at (0.0,\Hc) {ceiling};

% ---- robot-clearance reference (far left) ----------------------------
\draw[clr, |<->|] (0.55,0) -- (0.55,\Rz);
\node[clr, anchor=west, font=\scriptsize, align=left] at (0.62,0.5*\Rz)
  {$R^{}_{\max Z}$\\(robot)};

% ---- helper commands -------------------------------------------------
\newcommand{\colcap}[2]{%
  \node[lbl, anchor=south, align=center] at (#1,\Hc+0.12){#2};%
}

\newcommand{\verdict}[3]{%
  \fill[#2] (#1-\hw,-1.35) rectangle (#1+\hw,-0.75);
  \draw[#2!55!black] (#1-\hw,-1.35) rectangle (#1+\hw,-0.75);
  \node[white, font=\scriptsize] at (#1,-1.05) {#3};
  \draw[->, fsInk!70] (#1,-0.18) -- (#1,-0.70);%
}

% ============================ (A) open ================================
\def\xA{2.2}
\fill[free] (\xA-\hw,0) rectangle (\xA+\hw,\Hc);
\colcap{\xA}{(A) open}
\draw[brace, |-|] (\xA+\hw+0.12,0) -- (\xA+\hw+0.12,\Hc);
\node[fsFree!55!black, font=\scriptsize, anchor=west] 
  at (\xA+\hw+0.18,0.5*\Hc)
  {\textcolor{fsFree!60!black}{\checkmark}};
\verdict{\xA}{fsFree}{free}

% ====================== (B) low furniture =============================
\def\xB{4.0}
\fill[occ]  (\xB-\hw,0)   rectangle (\xB+\hw,0.8);
\node[white, font=\scriptsize] at (\xB,0.4) {furn.};
\fill[free] (\xB-\hw,0.8) rectangle (\xB+\hw,\Hc);
\colcap{\xB}{(B) low\\furniture}
\draw[brace, |-|] (\xB+\hw+0.12,0.8) -- (\xB+\hw+0.12,\Hc);
\node[fsFree!55!black, font=\scriptsize, anchor=west] 
  at (\xB+\hw+0.18,{0.5*(\Hc+0.8)})
  {\textcolor{fsFree!60!black}{\checkmark}};
\verdict{\xB}{fsFree}{free}

% ========================= (C) full wall ==============================
\def\xC{5.8}
\fill[occ] (\xC-\hw,0) rectangle (\xC+\hw,\Hc);
\node[white, rotate=90] at (\xC,0.5*\Hc) {wall};
\colcap{\xC}{(C) full-height\\wall}
\node[red!70!black, font=\scriptsize, anchor=west] 
  at (\xC+\hw+0.12,0.5*\Hc)
  {no run\,$\times$};
\verdict{\xC}{fsOcc}{obst.}

% ====================== (D) overhead beam =============================
\def\xD{7.6}
\fill[free] (\xD-\hw,0)   rectangle (\xD+\hw,2.3);
\fill[occ]  (\xD-\hw,2.3) rectangle (\xD+\hw,\Hc);
\node[white, font=\scriptsize] at (\xD,2.85) {beam};
\colcap{\xD}{(D) overhead\\beam}
\draw[brace, |-|] (\xD+\hw+0.12,0) -- (\xD+\hw+0.12,2.3);
\node[fsFree!55!black, font=\scriptsize, anchor=west] 
  at (\xD+\hw+0.18,1.15)
  {$\ge R^{}_{\max Z}$\,\textcolor{fsFree!60!black}{\checkmark}};
\verdict{\xD}{fsFree}{free}

% ---- resulting-layer label -------------------------------------------
% moved left to avoid overlapping the first verdict box
\node[lbl, anchor=east] at ({\xA-\hw-0.25},-1.05) {2D layer:};

\end{tikzpicture}%
}
\caption{2D free-space projection converts a 3D occupancy map into a 2D free-space layer~\cite{fredriksson2024voxel}. Each ground cell is classified by scanning its vertical voxel column for a free run with clearance at least $R^{}_{\max Z}$. Cells with sufficient clearance are marked \emph{free} (green), while blocked columns are marked \emph{obstacle} (gray). Open floor (A), low furniture (B), and overhead clearance (D) satisfy this condition, whereas a full-height wall (C) does not.}
\label{fig:free_space_projection}
\end{figure} 

\subsection{Geometric Perception}
\label{subsec:low_level_graph}
We first build the geometric substrate for room grounding. Posed RGB-D frames are back-projected into point clouds, integrated into an octree-based 3D occupancy map~\cite{hornung2013octomap}, and collapsed into a 2D free-space layer~\cite{fredriksson2024voxel}. The projection is purely geometric and represents traversable clearance rather than object semantics. For each ground-plane cell, the corresponding vertical voxel column is scanned to determine whether sufficient free space exists for the robot (\Cref{fig:free_space_projection}). This preserves architectural boundaries such as walls while suppressing low or overhead clutter, allowing the subsequent decomposition to follow the room layout rather than furniture-level structure.

For the reported experiments, the occupancy map uses a voxel resolution of $0.05$\,m and an $8$\,m sensor integration range. A ground cell is marked free only when its vertical voxel column contains a free gap of at least $R^{}_{\max Z}=0.3$\,m. The pipeline is driven by the ground-truth camera poses supplied with the dataset rather than by an odometry or SLAM frontend.

Semantic detections for the object layer are generated alongside the geometric mapping. At each frame, a closed-vocabulary YOLO instance-segmentation model~\cite{yolo26_ultralytics} produces 2D instance masks on the RGB image; each mask is back-projected into the global frame using the corresponding depth image, yielding a set of 3D detections $D^{}_t$ that feeds the object-node construction in \Cref{subsec:graph_maintenance}.

\subsection{Region Grounding}
\label{subsec:region_extraction}
We propose region grounding as the core novelty of this work. Rather than treating rooms as a cluster of navigable nodes or semantic labels, we anchor each room node to a single tracked free-space region: a polygonal decomposition of the 2D free-space layer that persists and updates as the robot explores. This gives each room an explicit, persistent geometric footprint defined by observed free space, making the room layer directly evaluable against room-scale geometric annotations.

\textbf{Free-space decomposition into regions. }
The 2D free-space layer is partitioned into polygonal regions using an incremental form of DUDE~\cite{fermin2017incremental}, which separates open space at natural constrictions such as doorways, narrow passages, and openings between adjacent areas.
Each update produces region polygons and adjacency relations, which define the spatial supports used to construct room nodes.

\textbf{Region tracking. }
Since decomposition can change as the map grows, regions are tracked across successive updates by associating each new region with the prior region with the greatest intersection-over-union (IoU). An association is accepted only when the IoU exceeds $0.20$, the centroid displacement is below $1.5$\,m, and the area ratio of the two regions falls within $[0.25,\,4.0]$. A region that fails to match is retained for up to $3$ consecutive updates before being discarded. A decomposition is rejected outright—and the last valid state is reused—when the total free area or region count drops by more than $15\%$ relative to the previous valid decomposition. These filters stabilize room identity against local mapping noise and transient reconstruction artifacts.

\textbf{Region-to-room anchoring. }
Each tracked region supports at most one room. During \textit{bootstrap}, the first valid robot pose is resolved to its containing region, and the first room inherits that region polygon as its footprint.
During \textit{active updates}, the region containing the current robot pose either refreshes its existing room node or anchors a new one if none exists for that region.
A \textit{maintenance} phase, executed every $0.5$\,Hz,
keeps the room layer globally consistent. For each stable tracked region, the system gathers entities whose geometry is assigned
to the region, such as object nodes, robot pose nodes, and navigation nodes. An unowned tracked region is promoted to a room anchor when this evidence set is non-empty, \eg when at least one such entity lies inside the region geometry (or within the configured boundary tolerance).
The maintenance phase also re-links room nodes to their current overlapping successor region when a tracked region is replaced and removes room nodes whose supporting regions have permanently disappeared.
This one-to-one region--room correspondence gives every room a stable identity and an explicit polygon; it also means room granularity inherits the decomposition, so a region that merges or splits physical rooms yields a correspondingly merged or split room node (\Cref{sec:qualitative}).

\subsection{Graph Construction}
\label{subsec:graph_maintenance}
After rooms obtain explicit polygons, the remaining 3DSG layers are assembled around them.
\textit{Agent} nodes are sampled from robot odometry and connected by temporal edges.
\textit{Object} nodes are built from the 3D detections $D^{}_t$ described in \Cref{subsec:low_level_graph}: detections of the same class whose centroids lie within $0.75$\,m are merged into a single object node, while unlabeled detections are kept distinct.
\textit{Navigation} nodes tile the 2D free-space layer into square blocks of $15\times15$ cells ($0.75$\,m at the $0.05$\,m map resolution); a block yields one navigation node when it contains at least $25$ free cells, and neighboring blocks are joined by traversability edges under $8$-connectivity.
Each agent, object, and navigation node is assigned to the room whose polygon contains its position, using a boundary tolerance of $0.5$\,m, and linked by a containment edge.
Room adjacency is added only when the corresponding supporting regions are adjacent and connected by a navigable crossing, so room-to-room edges represent traversable passages rather than geometric proximity alone. 
%The resulting graph is illustrated in~\Cref{fig:teaser_3dsg}.

% After rooms obtain explicit polygons, the remaining 3DSG layers are assembled
% around them. \textit{Agent} nodes are sampled from robot odometry and connected by temporal edges; \textit{object} nodes are produced by transforming semantic detections into a global frame and merging nearby detections of compatible classes; and \textit{navigation} nodes are derived from traversable free-space blocks. \IC{Specify how the semantic detections are obtained (which 2D/3D detector or segmentation model, closed- vs.\ open-vocabulary) and the exact merge rule for objects (distance threshold and class-compatibility criterion). Also state how navigation blocks are sampled (the table mentions a 15-cell stride / 8-connected graph. confirm and reference).} Each agent, object, and navigation node is assigned to the room whose polygon contains its position, using a small boundary tolerance, and linked by a containment edge.
% Room adjacency is added only when the corresponding supporting regions are adjacent and connected by a navigable crossing, so room-to-room edges represent traversable passages rather than geometric proximity alone. The resulting graph is illustrated in~\Cref{fig:teaser_3dsg}.

\begin{table*}[t]
\caption{Per-scene room-level results on 12 MP3D scenes (IoU threshold $0.15$, Hungarian assignment).
\textbf{O}: Ours; \textbf{H}: Hydra. \textbf{Bold}: better value per scene; ties not bolded.
\emph{GT}: ground-truth room count;
\emph{R}: recall; \emph{P}: precision; \emph{F1}: harmonic mean of R and P;
\emph{mIoU}: mean IoU over valid matches (excluded for Hydra on \textit{zsNo4HB9uLZ} where no valid match exists);
\emph{bF1}: boundary F1 at $0.25$\,m tolerance ($0.1$\,m sampling).}
\label{tab:main_results}
\centering
\begin{tabular*}{\textwidth}{@{\extracolsep{\fill}} l c rr rr rr rr rr}
\noalign{\hrule height 1.5pt}
\rule{0pt}{9pt}
  & & \multicolumn{2}{c}{\textbf{R}} &
      \multicolumn{2}{c}{\textbf{P}} &
      \multicolumn{2}{c}{\textbf{F1}} &
      \multicolumn{2}{c}{\textbf{mIoU}} &
      \multicolumn{2}{c}{\textbf{bF1}} \\
\cmidrule(lr){3-4}\cmidrule(lr){5-6}\cmidrule(lr){7-8}\cmidrule(lr){9-10}\cmidrule(lr){11-12}
\rule{0pt}{8pt}\textbf{Scene} & \textbf{GT} &
  \textbf{O} & \textbf{H} & \textbf{O} & \textbf{H} & \textbf{O} & \textbf{H} &
  \textbf{O} & \textbf{H} & \textbf{O} & \textbf{H} \\
\noalign{\hrule height 0.4pt}
17DRP5sb8fy & 10 & \textbf{.400} & .100 & .500          & .500          & \textbf{.444} & .167          & \textbf{.313} & .303          & \textbf{.183} & .000 \\
2t7WUuJeko7 &  6 & \textbf{.667} & .167 & .800          & \textbf{1.00} & \textbf{.727} & .286          & \textbf{.333} & .242          & \textbf{.125} & .058 \\
8WUmhLawc2A & 18 & \textbf{.333} & .167 & .300          & \textbf{.500} & \textbf{.316} & .250          & .307          & \textbf{.333} & \textbf{.132} & .052 \\
HxpKQynjfin &  8 & .125          & .125 & .500          & \textbf{1.00} & .200          & \textbf{.222} & \textbf{.751} & .248          & \textbf{.533} & .193 \\
JeFG25nYj2p & 22 & \textbf{.364} & .091 & .500          & \textbf{.667} & \textbf{.421} & .160          & .322          & \textbf{.369} & .204          & \textbf{.266} \\
jh4fc5c5qoQ & 16 & \textbf{.250} & .125 & .571          & \textbf{.667} & \textbf{.348} & .211          & .293          & \textbf{.324} & .185          & \textbf{.215} \\
Pm6F8kyY3z2 &  4 & \textbf{.500} & .250 & \textbf{.667} & .500          & \textbf{.571} & .333          & .266          & \textbf{.473} & \textbf{.282} & .033 \\
RPmz2sHmrrY &  8 & \textbf{.500} & .125 & .500          & \textbf{1.00} & \textbf{.500} & .222          & \textbf{.432} & .399          & \textbf{.403} & .240 \\
x8F5xyUWy9e & 15 & \textbf{.467} & .133 & .583          & \textbf{1.00} & \textbf{.519} & .235          & \textbf{.299} & .267          & \textbf{.146} & .134 \\
YVUC4YcDtcY & 11 & .273          & .273 & .500          & \textbf{1.00} & .353          & \textbf{.429} & \textbf{.263} & .200          & \textbf{.083} & .033 \\
Z6MFQCViBuw & 22 & \textbf{.318} & .273 & .412          & \textbf{.600} & .359          & \textbf{.375} & \textbf{.447} & .392          & \textbf{.183} & .129 \\
zsNo4HB9uLZ & 23 & \textbf{.348} & .000 & \textbf{.381} & .000          & \textbf{.364} & .000          & \textbf{.344} & {--}           & \textbf{.234} & .000 \\
\noalign{\hrule height 0.4pt}
\textbf{Average} & \textbf{13.6} &
  \textbf{.379} & .152 & .518 & \textbf{.703} & \textbf{.427} & .241 &
  \textbf{.364} & .323 & \textbf{.224} & .113 \\
\noalign{\hrule height 1.5pt}
\end{tabular*}
\end{table*}

\section{Evaluation}
\label{sec:evaluation}

We evaluate the generated room layer as 2D room-instance polygons matched to the annotated rooms of each scene. The evaluation focuses on three aspects: \emph{instance recovery} (does the room layer recover the room instances of the scene?), \emph{spatial
coverage} (how much of the annotated room space is covered by matched rooms?),
and \emph{boundary fidelity} (how closely do room footprints follow the true room
boundaries?).

\subsection{Dataset and Setup}
\label{sec:eval_setup}

We evaluate 12 Matterport3D (MP3D) scenes~\cite{chang2017matterport3d} spanning between 4--23 ground-truth rooms each. Ground-truth room polygons are obtained by projecting the MP3D per-room surface annotations onto the ground plane and extracting one footprint per annotated room.  Each scene is replayed as an RGB-D trajectory so that the pipeline observes it incrementally. We compare against Hydra~\cite{hughes2024foundations} with its default configuration on the same scenes.
Hydra is the only system
whose room layer can be readily converted into comparable room polygons; others either do not expose room footprints, rely on point-cloud or wall-plane representations that require non-trivial post-processing to obtain a footprint, or are not available in a form suitable for direct evaluation. We therefore use it as the sole direct baseline. Both methods are evaluated as 2D room-instance polygon generators.

\subsection{Metrics and Scope}
\label{sec:metrics}

We evaluate room polygons using 2D room-segmentation metrics based on polygon overlap~\cite{luperto2022robust, bao2025topology}. Predicted and ground-truth polygons are first centroid-aligned and matched one-to-one with Hungarian assignment over intersection-over-union (IoU). A match is valid when IoU $\geq 0.15$, a permissive threshold chosen to account for partial MP3D trajectory coverage and to measure approximate room-instance association rather than strict wall-level segmentation.

We report recall, precision, and F1 to measure room-instance recovery: recall is the fraction of ground-truth rooms matched by a prediction, precision is the fraction of predicted rooms matched to ground truth, and F1 summarizes their balance. For valid matches, we report matched-only mean IoU (mIoU) to measure footprint overlap and boundary F1 (bF1) to measure boundary alignment using a 0.25~m tolerance and 0.1~m boundary sampling. We also report over-segmentation (oS), the fraction of ground-truth rooms split across multiple predictions, and under-segmentation (uS), the fraction of predicted rooms spanning multiple ground-truth rooms.

\subsection{Quantitative Results}
\label{sec:results}

\Cref{tab:main_results} shows a clear improvement in room-instance recovery. Occupancy-grounded anchoring roughly doubles recall over Hydra (0.152 to 0.379), achieves higher F1 in 9 of 12 scenes, and recovers nearly five ground-truth rooms per scene on average, compared with fewer than two for Hydra, but at lower precision. The precision drop stems  from unmatched room nodes whose region polygons are too poorly bounded to satisfy the one-to-one IoU criterion (\Cref{sec:eval_setup}), so the dominant errors are boundary fidelity and instance separation, not coverage. Overall, the results expose a coverage-precision trade-off that recall and F1 capture, but footprint-overlap scores can obscure when several rooms collapse into one single matched instance.

\subsection{Qualitative Analysis}
\label{sec:qualitative}

\begin{figure*}[t]
    \centering
    \includegraphics[width=0.32\textwidth,height=0.18\textheight,keepaspectratio]{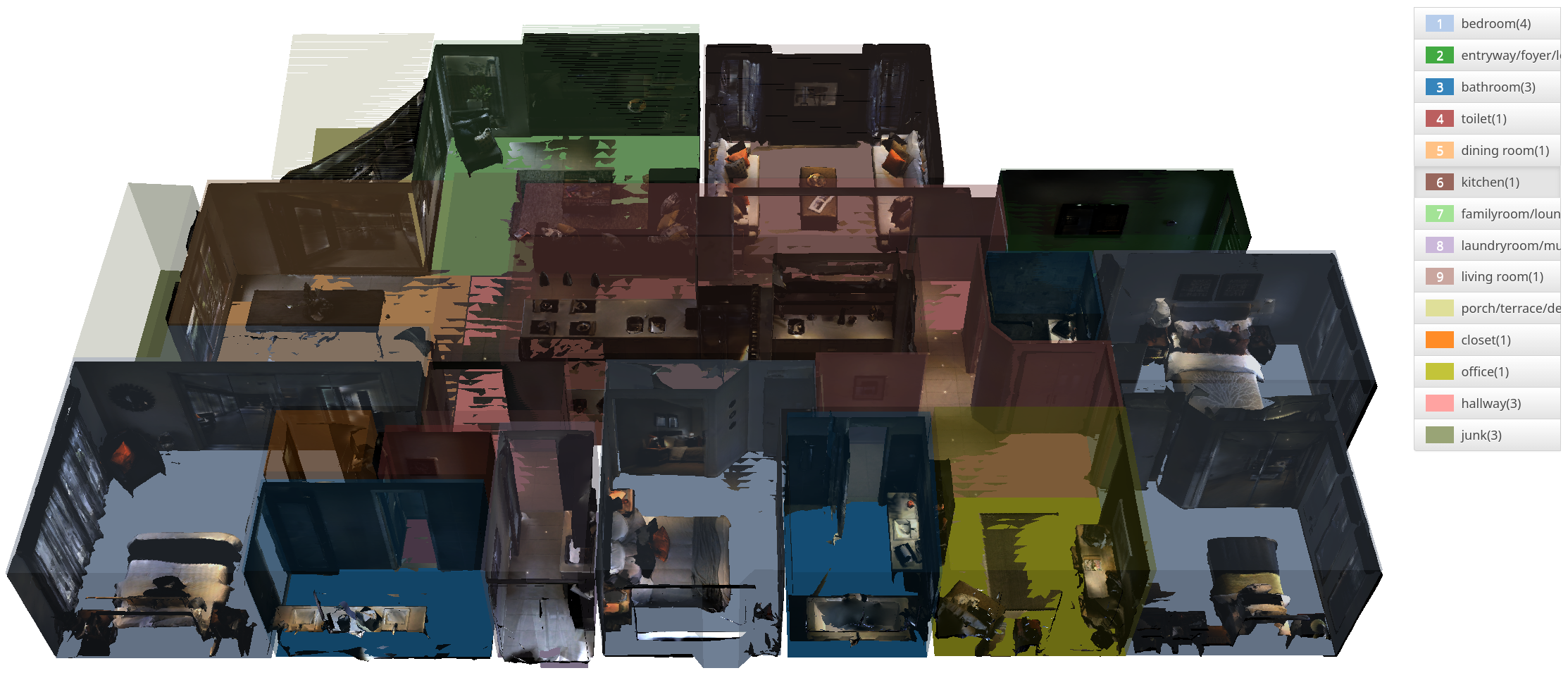}\hfill
    \includegraphics[width=0.32\textwidth,height=0.18\textheight,keepaspectratio]{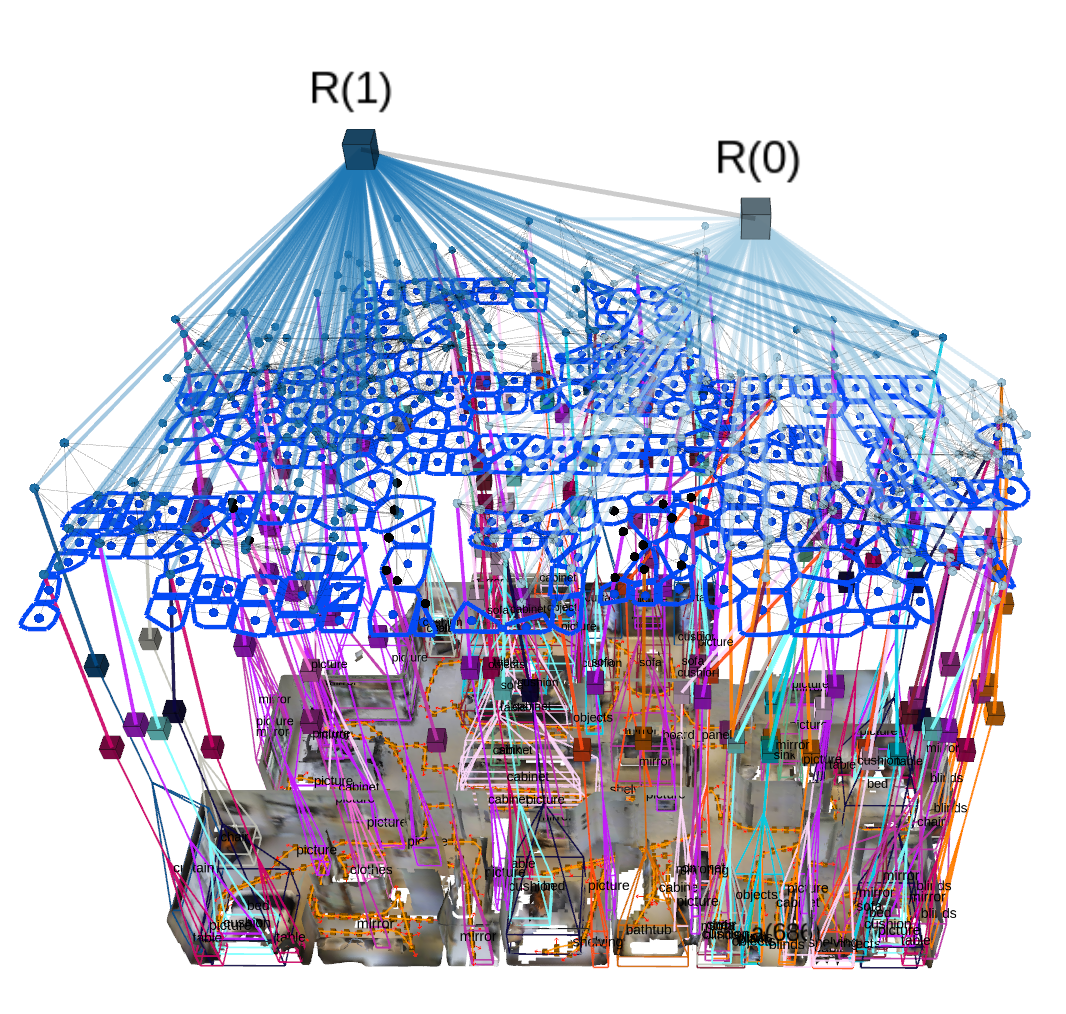}\hfill
    \includegraphics[width=0.32\textwidth,height=0.18\textheight,keepaspectratio]{figures/MINE/zsNo4HB9uLZ.png}
    \caption{Representative qualitative results. Ground-truth images show MP3D room annotations; Hydra and Ours show the generated scene graph with room nodes and associated spatial support.}
    \label{fig:qualitative}
\end{figure*}

Although both methods face a precision--recall trade-off, the difference lies in where and why collapse occurs \Cref{fig:qualitative}). In Hydra, collapse arises from navigational nodes clustering: dense connectivity merges distinct rooms, yielding only two nodes in the 23-room scene and one in a 6-room scene. Its high precision, therefore, reflects a conservative, underpopulated room layer rather than accurate room segmentation.

In our pipeline, collapse is local and originates in the free-space decomposition. Non-convex layouts or narrow connections can cause the decomposer to merge adjacent rooms into one support region, so anchoring cannot recover the missing instances. When rooms are separated cleanly, one-room-per-region anchoring improves recall at the cost of occasional spurious nodes. The practical implication is that recall is bounded by decomposition granularity: recovering more rooms requires finer decomposition, not a different anchoring strategy.

A second failure mode is boundary degradation from input-map errors. Reconstruction gaps can introduce spurious free cells that the decomposer treats as navigable space, merging geometrically separate areas into one region and propagating the error to the room layer. The implication here is that room quality is ultimately bounded by map quality: errors in reconstruction flow through free-space projection and region decomposition before reaching the room layer, so no anchoring strategy can recover what the input map loses.

\begin{table}[t]
    \caption{Ablation of the DUDE threshold $d^{}_{\mathrm{thr}}$ on three MP3D scenes (macro-averages over \textit{17DRP5sb8fy}, \textit{RPmz2sHmrrY}, and \textit{zsNo4HB9uLZ}).
    \emph{Pred}: predicted room nodes; \emph{M}: valid IoU matches;
    \emph{R}/\emph{P}/\emph{F1}: recall, precision, F1 at IoU threshold $0.15$;
    \emph{mIoU$_{\mathrm{GT}}$}: GT-penalized mean IoU (unmatched GT rooms contribute zero);
    \emph{Severe}: scenes flagged for severe under-segmentation.}
    \label{tab:ablation_dude_decomposition_quantitative}
    \centering
    \resizebox{\columnwidth}{!}{%
        \begin{tabular}{l r r r r r r r r}
        \noalign{\hrule height 1.5pt}
        \rule{0pt}{9pt}\textbf{Param.} & \textbf{Pred} & \textbf{M} & \textbf{R} & \textbf{P} &
        \textbf{F1} & \textbf{mIoU$_{\mathrm{GT}}$} & \textbf{bF1} &
        \textbf{Severe} \\
        \noalign{\hrule height 0.4pt}
        $1.5$ & 12.33 & 5.33 & .416 & .460 & .436 & .154 & .273 & 0/3 \\
        $3.0$ &  6.00 & 1.67 & .118 & .278 & .163 & .035 & .183 & 3/3 \\
        $6.0$ &  2.00 & 0.33 & .014 & .083 & .025 & .005 & .069 & 3/3 \\
        \noalign{\hrule height 1.5pt}
        \end{tabular}%
    }
\end{table}

\subsection{Ablation Study}
\label{subsec:ablation_dude_decomposition}

We ablate the DUDE threshold $d^{}_{\mathrm{thr}}$, which controls the granularity
of the free-space decomposition and the number of region supports available for
rooms. We test $d^{}_{\mathrm{thr}} \in \{1.5, 3.0, 6.0\}$ on
\textit{17DRP5sb8fy}, \textit{RPmz2sHmrrY}, and \textit{zsNo4HB9uLZ}; the main
results use $d^{}_{\mathrm{thr}}=1.5$.

\Cref{tab:ablation_dude_decomposition_quantitative} shows that increasing
$d^{}_{\mathrm{thr}}$ makes the decomposition progressively coarser. The average
number of predicted rooms decreases from 12.33 to 6.00 and then to 2.00, while
recall drops from 0.416 to 0.118 and 0.014, and F1 from 0.436 to 0.163 and
0.025. 
% The qualitative results in \Cref{tab:ablation_dude_decomposition_qualitative} show the same trend:
This reflects a fine-to-coarse behavior: $d^{}_{\mathrm{thr}}=1.5$ separates more room-scale areas, $d^{}_{\mathrm{thr}}=3.0$
merges some neighboring rooms, and $d^{}_{\mathrm{thr}}=6.0$ groups most free space
into a few large supports. Thus, increasing $d^{}_{\mathrm{thr}}$ trades
room-instance recall for coarser free-space connectivity.

\begin{table}[t]
\caption{Mean per-update runtime of each pipeline stage.}
\label{tab:runtime}
\centering
\resizebox{\columnwidth}{!}{%
\begin{tabular}{l r r}
\noalign{\hrule height 1.5pt}
\rule{0pt}{9pt}\textbf{Stage} & \textbf{Mean (ms)} & \textbf{Std (ms)} \\
\noalign{\hrule height 0.4pt}
Point-cloud generation & 0.34 & 0.06 \\
3D occupancy integration~\cite{hornung2013octomap} & 358.15 & 169.49 \\
2D free-space projection~\cite{fredriksson2024voxel} & 12.5 & 111.02 \\
DUDE decomposition~\cite{fermin2017incremental} & 2.5 & 7.07 \\
Region tracking & 0.04 & 0.02 \\
Entity assignment \& graph assembly & 1209.30 & 865.42 \\
\noalign{\hrule height 0.4pt}
\textbf{Total per update} & \textbf{1582.83} & \textbf{1053.08} \\
\noalign{\hrule height 1.5pt}
\end{tabular}%
}
\end{table}

\subsection{Runtime}
\label{sec:runtime}
Because the pipeline is incremental, room supports are maintained online as the map grows rather than recomputed from scratch. \Cref{tab:runtime} reports the mean per-update cost of each stage measured on a desktop with an AMD Ryzen 9 9900X, 64~GB RAM, GeForce RTX 5070 Ti, and running ROS~2 Humble; scene-graph output at $1$\,Hz average.

\section{Conclusion}
\label{sec:conclusion}

We presented a pipeline that exposes the room layer of a hierarchical 3DSG in a directly comparable form. Each \textit{room} node is anchored to a tracked, occupancy-derived free-space region and exported as an explicit polygon, enabling standard room-segmentation metrics to be applied to 3DSG room layers.

On 12 Matterport3D scenes, occupancy-grounded anchoring recovered more room instances than Hydra, increasing recall from 0.152 to 0.379 and F1 from 0.241 to 0.427. Hydra achieved higher precision per predicted room (0.703 vs.\ 0.518), largely because it predicted few rooms and under-segmented most scenes. These results should therefore be read as a coverage--precision trade-off rather than a general ranking. More than half of the ground-truth rooms remain unmatched on average, and boundary fidelity is limited by errors inherited from the free-space layer, as shown by the \textit{x8F5xyUWy9e} case. Room-instance recovery and boundary-accurate room footprints therefore remain open problems for hierarchical 3DSGs.

\section*{Acknowledgment}
The authors acknowledge the use of GPT-5.5~\cite{openai2026gpt55} for
improving readability across all sections and for documentation and code
readability in the project. The authors also acknowledge the financial support of the Finnish Ministry of Education and Culture through the Intelligent Work Machines Doctoral Education Pilot Program (IWM VN/3137/2024-OKM-4) and the Finnish Cultural Foundation.

\bibliographystyle{IEEEtran}
\bibliography{bibliography}

\vfill

\end{document}